\documentclass[sigconf,screen=true,authorversion]{acmart}

\AtBeginDocument{%
	\providecommand\BibTeX{{%
			\normalfont B\kern-0.5em{\scshape i\kern-0.25em b}\kern-0.8em\TeX}}}

\setcopyright{acmcopyright}
\copyrightyear{2022}
\acmYear{2022}
\setcopyright{acmcopyright}\acmConference[ACM SE '22]{2022 ACM Southeast Conference}{April 18--20, 2022}{Virtual Event, USA}
\acmBooktitle{2022 ACM Southeast Conference (ACM SE '22), April 18--20, 2022, Virtual Event, USA}
\acmPrice{15.00}
\acmDOI{10.1145/3476883.3520216}
\acmISBN{978-1-4503-8697-5/22/04}

\begin{document}
	\pagestyle{empty}
	
	\title{Superpixel-based Knowledge Infusion in Deep Neural Networks for Image Classification}
	

	\author{Gunjan Chhablani}
	\affiliation{%
		\department{Dept. of CSIS and APPCAIR}
		\institution{BITS Pilani, Goa Campus}
		\city{Zuarinagar}
		\country{IN}
	}
	
	\author{Abheesht Sharma}
	\affiliation{%
		\department{Dept. of CSIS and APPCAIR}
		\institution{BITS Pilani, Goa Campus}
		\city{Zuarinagar}
		\country{IN}
	}
	
	\author{Harshit Pandey}
	\affiliation{%
		\department{Dept. of CS}
		\institution{Savitribai Phule Pune University}
		\city{Pune}
		\country{IN}
	}
	
	\author{Tirtharaj Dash}
	\affiliation{%
		\department{Dept. of CSIS and APPCAIR}
		\institution{BITS Pilani, Goa Campus}
		\city{Zuarinagar}
		\country{IN}
	}

	\renewcommand{\shortauthors}{Chhablani \textit{et al.}}
	
	\begin{abstract}
		Superpixels are higher-order perceptual groups of pixels in an image, often carrying much more information than the raw pixels. 
		There is an inherent relational structure to the relationship among different superpixels of an image
		such as adjacent superpixels are neighbours
		of each other. Our interest here is to treat
		these relative positions of various superpixels 
		as relational information of an image.
		This relational information can convey higher-order spatial information about the image, such as the relationship between superpixels representing two eyes in an image of a cat. That is, two eyes are placed adjacent to 
		each other in a straight line or the mouth is below
		the nose.
		Our motive in this paper is to assist computer vision models, specifically those based on Deep Neural Networks (DNNs), by incorporating this higher-order information from superpixels.
		We construct a hybrid model that leverages (a) Convolutional Neural Network (CNN) to deal with spatial information in an image and (b) Graph Neural Network (GNN) to deal with relational superpixel information in the image.
		The proposed model is learned using a generic hybrid loss function. 
		Our experiments are extensive, and we evaluate the predictive performance of our proposed hybrid vision model on seven different image classification datasets from a variety of 
		domains such as
		digit and object recognition, biometrics,
		medical imaging.
		The results demonstrate that the relational superpixel information processed by a GNN can improve the performance of a standard CNN-based vision system.
	\end{abstract}
	
	
	\ccsdesc[500]{Computing methodologies~Neural networks}
	\ccsdesc[500]{Computing methodologies~Computer vision representations}

	\keywords{Knowledge-Infused Learning, Graph Neural Networks, Convolutional Neural Networks, Superpixels, SLIC}
	
	
	\maketitle
	
	\section{Introduction}
	\label{sec:introduction}
	
	In the ever-burgeoning field of Deep Learning, the task of image classification and recognition has taken centre stage, chiefly after the introduction of the ILSVRC Challenge\footnote{\url{https://image-net.org/challenges/LSVRC/}}.
	There have been significant architectural innovations for Convolutional Neural Networks (CNNs).
	In the last few years, the core approach of basic convolution has been adopted to graph-structured data via the introduction of graph neural networks (GNNs), a term coined by Scarselli~\textit{et al.}~\cite{scarselli2008graph}. 
	
	A graph is a representation of binary relations. 
	GNNs can utilise this relational information during their learning. The learning process includes information flow using the \textsf{AGGREGATE} operation. GNNs can construct a graph representation useful for the supervised learning task at hand (e.g., graph classification or regression).
	These binary relations can be seen in the data instances. In graphs, binary relations are represented as edges.
	For instance, a graph of countries may have edges between countries that share borders; a graph of cities may have an edge between two cities if they share a bus service; a network of papers, where papers that cite each other are related.  
	In the case of tasks involving images, binary relations can be easily seen at the level of `image superpixels'. Superpixels are higher-order perceptual groups of pixels in an image. Superpixels group an image into meaningful atomic regions that can be used to replace the rigid structure of the pixel grid in images. They often convey much more information than low-level raw pixels and share some common characteristics such as intensity levels~\cite{ren2003learning}. 
	
\begin{figure}[!htb]
		\centering
		\begin{minipage}{0.5\linewidth}
			\centering
			\includegraphics[width=0.8\textwidth]{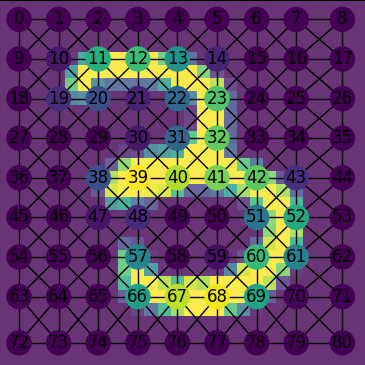}
		\end{minipage}%
		\begin{minipage}{0.5\linewidth}
			\centering
			\includegraphics[width=0.8\textwidth]{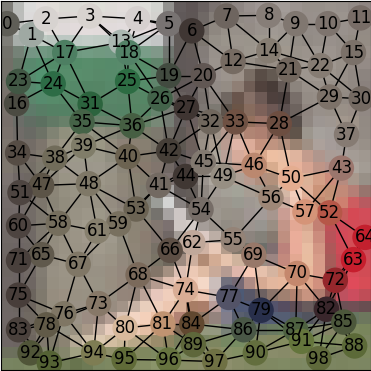}
		\end{minipage}
		\caption{Superpixels for the digit `3' image (left), a boy reading a book (right)}
		\label{fig:exsuperpix}
	\end{figure}

	Figure~\ref{fig:exsuperpix} shows the super-pixellated version of two raw images. 
	One can observe that the superpixels share binary relations with their neighbours, resulting in a graph structure. 
	Our present work treats the relational information conveyed by the superpixels as higher-order spatial information about an image and provides this information to aid in image classification tasks. 
	It has been observed that incorporating (domain) knowledge can significantly enhance the performance of deep learning models~\cite{dash2021vegnn}, even in problems where the amount of available data is low~\cite{dash2021incorporating}.
	Although the higher-level spatial information encoded in 
	the image superpixels cannot directly be called ``domain knowledge'' the neighbourhood structure represented
	by the connections among the superpixels does loosely 
	convey some form of relational or domain information.
	Our intention in this work is to investigate
	a methodology of infusing such knowledge into a 
	deep-learning pipeline for image classification problems.
	
	In this work, we leverage spatial information from the image and infuse knowledge in the form of binary relations procured from the superpixel graph representation of the image. 
	CNN filters tend to learn parameters based on pixel-level information. 
	We hypothesise that fusing such superpixel-level information can provide a higher-level understanding of the image and aid a CNN in classification tasks. 
	Specifically, we treat the graph resulting from the image superpixels as an input to a GNN and learn a CNN-based vision system together with the GNN. 
	The coupled hybrid CNN+GNN system is expected to be knowledge-rich, both at the level of raw pixels and superpixels. 
	
	\subsubsection*{Major contributions} 
	Overall, the major contributions of our paper are as follows:
	
	\begin{enumerate}
		\item Treating the superpixel representation of an image as a graph and allowing a GNN to extract higher-level domain information about an image;
		\item We construct an image classification model by coupling the GNN with a CNN-based baseline; 
		\item We conduct a series of empirical evaluations of our coupled CNN+GNN hybrid system on four different popular image classification benchmarks and three case studies.
	\end{enumerate}

	The rest of the paper is organised as follows. 
	In Section~\ref{sec:integrated_model}, we provide details of our methodology, including a new loss function suitable for learning the hybrid model. 
	Section~\ref{sec:emp_eval} provides details of our experiments. 
	Section~\ref{sec:related_work} provides a brief description of related work. 
	Section~\ref{sec:conclusion} concludes the paper.
	
	\section{Superpixels Integrated Vision Model}
	\label{sec:integrated_model}
	
	
	\subsection{Superpixel Graph Construction}
	\label{sec:superpixel_construction}
	
	Simple Linear Iterative Clustering (SLIC)~\cite{achanta2012slic} is an easy-to-use and straightforward algorithm. SLIC generates
	superpixels by clustering pixels of an image
	based on their colour similarity and
	proximity in the image plane.
	As a black-box procedure, as is the case in our present study, 
	SLIC takes an approximate number of superpixels and the input image as its input and outputs a segmented image. 
	We use SLIC to construct the segmented version of the input image: each segmented patch in the image is now a superpixel. 
	We then treat the centroid of every superpixel as a node in the graph. We link these nodes by building a radius graph~\cite{bentley1977complexity}.  
	For every pair of nodes, we form an edge between them if and only if the Euclidean distance between them is less than a pre-ordained radius, $r \in \mathbb{R}$. 
	Mathematically, a radius graph is defined as $G=(V, E)$ such that:
	
	\begin{align*}
		V &= \{x_i: x_i ~\text{is a superpixel}\}, \\
		E &= \{\{x_i,x_j\}: x_{i,j} \in V, x_i \neq x_j, ||x_i - _j|| \leq r\}
	\end{align*}
	
	\subsection{The Coupled CNN+GNN Model}
	
\begin{figure*}[!htb]
		\centering
		\includegraphics[width=0.8\linewidth]{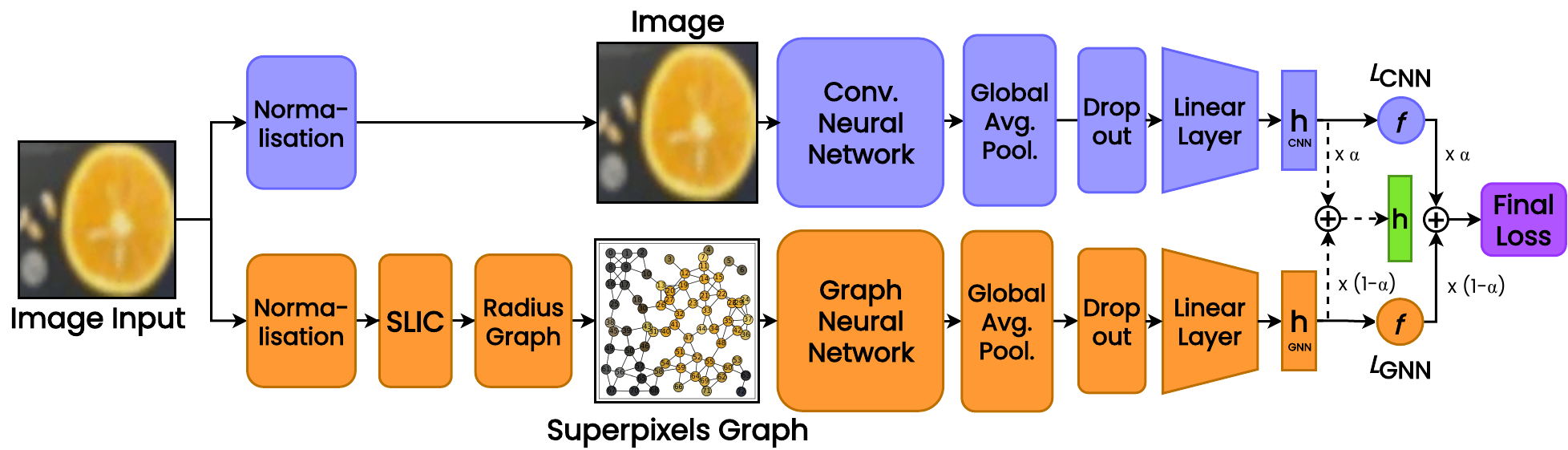}
		\caption{Block diagram of the proposed CNN+GNN model. The hybrid model consists of two backbones: a CNN backbone and a GNN backbone. The input to the CNN backbone is an image, and the input to the GNN backbone is the radius graph (obtained from the superpixel representation) corresponding to the image. The CNN block consists of multiple sub-blocks (layers) of Convolution (C) and Max. Pooling (P) blocks interleaving each other, e.g. CPCPCP. Similarly, there are multiple Graph Convolution blocks (graph attention layers, as in GAT) in the GNN block. The Global Mean Pool blocks read out the final representation from the CNN and GNN. The dropout block (optional) is used to deal with noise in representation and produce a robust representation. $h$ denotes the logits, $\alpha$ is a hyperparameter described later. $\oplus$ is the addition operation.}
		\label{fig:cnngnnmodel}
	\end{figure*}

	Our main motive in this work is to learn jointly from the image and its superpixel representation. This results in a complementary combination of a vision model (CNN) and a relational model (GNN).
	The CNN takes care of feature extraction (low-level features such as edges in initial layers to complex objects in the latter layers). In contrast, the graph-based model takes the superpixel-based radius graph and extracts relational information about the image that can act as domain knowledge. 
	Fig.~\ref{fig:cnngnnmodel} illustrates this hybrid combination.
	
	Regardless of the detailed architectural specifics, the goal is to construct a representation of the input (image) that consists of both the spatial information (extracted by the CNN) and the domain information (extracted by the GNN). 
	This representation can then be used to construct a feedforward fully-connected neural network that outputs a class label for the input image. 
	In order to train the coupled model in an end-to-end fashion and utilise the information from the combined representation adequately, we propose a simple hybrid loss as follows:
	\begin{equation}
	\mathcal{L}_{CG} = \alpha \mathcal{L}_{C} + (1-\alpha) \mathcal{L}_{G}
	\end{equation}
	Here $\mathcal{L}_{C}$ and $\mathcal{L}_{G}$ denote 
	the cross-entropy loss for CNN and GNN, respectively.
	The parameter $\alpha$ determines the relative importance accorded to the two models during training, i.e., a value $\alpha=0.5$ would mean that both the raw spatial information and the domain information are equally important. 
	For our construction, we treat this as a tunable hyperparameter.
	
	For inference on the trained model (i.e., prediction), we construct hybrid logits based on the logits computed by the CNN backbone and GNN backbone for an input image as follows:
	\begin{equation}
	\hat{y} = \arg \max_{j} (\alpha h_{C} + (1-\alpha) h_{G})
	\end{equation}
	where $h_{C}$ and $h_{G}$ are the logits from CNN and GNN respectively.
	

	\section{Empirical Evaluation}
	\label{sec:emp_eval}
	
	\subsection{Aims}
	\label{ssec:aims}
	This study aims to integrate superpixel-level domain knowledge with vision systems, specifically those based on CNNs. Our empirical experiments attempt to answer the following research questions: 
	\textbf{RQ1}: Can GNNs construct rich relational representations from superpixels?;
	\textbf{RQ2}: Can GNN-constructed knowledge improve
	CNN-based vision systems?
	
	\subsection{Data}
	\label{ssec:data}
	We test our hypothesis on a range of datasets: one dataset for handwriting recognition, one from the fashion and clothing domain, two from the object recognition domain, two datasets from the domain of biometrics, and one dataset from the domain of medical diagnosis.
	We briefly describe these datasets below. A summary of the dataset is provided in Fig.~\ref{tab:dataset} showing the number of classes, number of instances in training and testing set of each dataset.
	
	\subsubsection{MNIST} 
	MNIST~\cite{lecun1998mnist} is one of the most popular image classification datasets. It is a database of 28x28-sized grayscale images of handwritten digits, and the task is to identify the digit in the image. MNIST has 10 classes, one for every digit.
	
	\subsubsection{FMNIST} 
	Fashion-MNIST (FMNIST)~\cite{xiao2017fashion} is a sister dataset of MNIST, consisting of fashion and clothing items. This dataset was collated from Zalando's database of article images. Each image is a 28x28 grayscale image. The dataset has 10 classes, one for every fashion article type.
	
	\subsubsection{CIFAR-10} 
	CIFAR-10~\cite{krizhevsky2010cifar10} is an established object recognition dataset, consisting of 32x32 RGB images. It has ten different classes such as aeroplane, automobile, bird, cat, dog. There are 6000 instances
	for each class.
	
	\subsubsection{CIFAR-100} 
	CIFAR-100~\cite{krizhevsky2009cifar100} is another object recognition dataset, with 100 classes and RGB images, each of size 32x32. This is like the
	CIFAR-10 dataset, and there are 600 instances
	for each class.
	
	\subsubsection{LFW} 
	The Labelled Faces in the Wild (LFW~\cite{huang2007lfw,huang2007unsupervised}) dataset is a face recognition dataset. Every image is an RGB image of size 250x250. We do not consider all faces in the dataset; we discard faces with less than 20 faces in order to maintain the class balance. After discarding such faces, we have 62 classes.
	
	\subsubsection{SOCOFing} 
	The Sokoto Coventry Fingerprint~\cite{shehu2018sokoto} dataset is made up of 6,000 fingerprint images from 600 African subjects. All images are grayscale and have a resolution of 96x103. The task is to identify the person to which the given fingerprint sample belongs.
	
	\subsubsection{COVID-19 Radiography} 
	The COVID-19 Radiography Dataset~\cite{chowdhury2020can,rahman2021covid19} consists of chest X-ray images for patients with COVID-19, viral pneumonia, lung opacity as well as normal patients, i.e., four classes. Each image is a 299x299 grayscale image. This dataset has been
	one of the most popular datasets in machine learning
	studies involving COVID-19 disease.

	\begin{figure}[!htb]
		\centering
		\begin{tabular}{c|c|c|c}
			\hline
			\textbf{Dataset}  & \textbf{NumClasses} & \textbf{Trainset-size} & \textbf{Testset-size} \\ \hline
			MNIST & 10 & 60K & 10K \\
			FMNIST & 10 & 60K  & 10K \\
			CIFAR-10 & 10 & 50K  & 10K \\
			CIFAR-100 & 100 & 50K  & 10K \\
			LFW & 62 & 2.4K & 0.6K \\
			SOCOFing & 600 & 49K & 6K \\
			COVID-19 X-ray & 4 & 19K & 2K \\ \hline
		\end{tabular}
		\caption{Dataset Summary. The sizes shown are the
			approximate number of instances in the official training set and test set supplied along with each dataset.}
		\label{tab:dataset}
	\end{figure}

	\subsection{Algorithms and Machines}
	\label{sec:algomach}
	
	We use the method described in Sec.~\ref{sec:superpixel_construction} to construct a superpixel-based radius graph for each image in these datasets. For convenience, we refer to this procedure as $RadiusGraph$ that takes two inputs: a set of images and the number of superpixels ($n$), and returns a set of radius graphs obtained from the superpixel representation of the images.
	
	We use the PyTorch library for the implementation of CNN and PyTorch Geometric for GNN models. 
	We conduct all our experiments on OVHCloud's\footnote{\url{https://www.ovhcloud.com/en-ie/}} ``AI Training" platform with the following configurations:  32GB NVIDIA V100S GPU, 45GB main memory, 14 Intel Xeon 2.90GHz processors and a NVIDIA-DGX station with 32GB Tesla V100 GPU,256GB main memory, 40 Intel Xeon 2.20GHz processors.

	\subsection{Method}
	\label{ssec:method}
	
	Our method is straightforward. For each dataset ($D$),
	we determine the value of the number of superpixels 
	($n$) for the procedure $RadiusGraph$ based on
	visual analysis.
	Let $D$ be a dataset with a set of images $X$ and
	their corresponding class-labels $Y$. So,
	$D$ is written as $(X,Y)$.
	In the following steps, $D_{tr}$
	and $D_{te}$ denote the official train-set and
	test-set for a dataset $D$. These details can be obtained
	from the official sources for each dataset as referred
	in this paper. We outline the steps of our methodology
	below:
	
	\begin{enumerate}
		\item Let $X' = RadiusGraph(X,n)$
		\item $D' = (X', Y)$
		\item Let $D = D_{tr} \cup D_{te}$
		and $D' = D'_{tr} \cup D'_{te}$
		be the corresponding train-test splits
		for the image dataset ($D$) and the 
		superpixel graph dataset ($D'$)
		\item Construct a CNN model using $D_{tr}$: denote as $M_{c}$
		\item Construct a coupled CNN+GNN model
		using $D_{tr} \cup D'_{tr}$: denote as
		$M_{cg}$
		\item Let $P_{c}$ be the predictive performance of $M_{c}$ on $D_{te}$
		\item Let $P_{cg}$ be the predictive performance of $M_{cg}$ on $D_{te}$
		\item Compare $P_{c}$ and $P_{cg}$
	\end{enumerate}

	\noindent
	The following details pertaining to the above steps are relevant: 
	\begin{itemize}
		\item We use $n=75$ for MNIST and FMNIST, $n=100$ for CIFAR, LFW and SOCOFing and $n=200$ for COVID to construct superpixel graphs using $RadiusGraph$; the number of radial neighbours is set to $5$ for MNIST, FMNIST and CIFAR datasets, $27$ for COVID, $10$ for LFW, and $15$ for SOCOFing. This was decided using graph visualisation;
		
		\item The node feature-vector for the radius graphs are: the normalised pixel value and the location of centroid in the image, resulting in 3-length feature-vector for MNIST and FMNIST, and 5-length feature-vector for CIFAR, COVID and LFW;
		
		\item We use 90:10 split on $D_{tr}$ (correspondingly, $D'_{tr}$) to obtain a validation set useful for hyperparameter tuning;
		
		\item We use the AdamW optimiser~\cite{loshchilov2018decoupled} for training all our models;
		
		\item The hyperparameters are obtained by a sweep across grids such as batch-size: $\{128,256\}$, learning rate: \{1e-3,1e-4,1e-5\}, weight-decay parameter: $\{0,0.001\}$ for all the datasets; 
		
		\item CNN structure is built with 3-blocks; each block consists of a convolutional layer, a batch-norm layer, ReLU activation and a max-pool layer. The number of channels in each convolution layer is 32, 64 and 64, respectively;
		
		\item The coupled CNN+GNN structure consists of the same CNN backbone as mentioned above; and a GNN structure consisting of three graph-attention layers (GAT) with 32, 64 and 64 channels, respectively, and the number of heads is set to 1;
		
		\item To determine the optimal value of $\alpha$, we run a grid search over multiple values in the range $(0,1)$. $\alpha = 0.75$ was found to be the optimal value for our datasets. However, as 
		mentioned earlier, it is advised to use $\alpha$ as a
		trainable hyperparameter for future works.
		
		\item We use accuracy as the metric to measure the predictive performance of the models on $D_{te}$.
	\end{itemize}

	
	
	\subsection{Results}
	\label{ssec:results}
	
	In all our experiments, the primary intention is to show whether relational information provided via a superpixel graph is able to aid in improving the performance of a CNN-based model. 
	One should note that the primary backbone of the CNN in the standalone model (CNN) and that in the coupled model remains the same to draw any meaningful conclusion.
	The principal results of our experiments are reported in Fig.~\ref{fig:results}.
	The results demonstrate that the higher-level domain knowledge extracted by a GNN from the superpixel graph is able to substantially boost the predictive performance of CNNs in a hybrid-learning setting in some cases.
	
	\begin{figure}[!htb]
		\centering
		\begin{tabular}{c|c|c}
			\hline
			\textbf{Dataset}     &  \textbf{CNN}       &      \textbf{CNN+GNN} \\
			\hline
			MNIST       &  \textbf{99.30}     &      99.21  \\
			FMNIST      &  \textbf{91.65}     &      91.50  \\
			CIFAR-10    &  \textbf{77.80}     &      76.81  \\
			CIFAR-100   &  42.88     &      \textbf{46.79}  \\ \hline
			LFW  &  60.83   &  \textbf{66.12}   \\
			SOCOFing &   65.68     &   \textbf{93.58}   \\ \hline
			COVID Chest X-ray &   89.09     &   \textbf{91.01}  \\
			\hline
		\end{tabular}
		\caption{Predictive performance (Accuracy) of CNN and CNN+GNN on all datasets}
		\label{fig:results}
	\end{figure}
	
	Readers familiar with deep neural networks would
	agree that networks with complex structures and
	a large number of parameters tend to be more accurate
	than those with a lesser number of parameters with
	a simple structure. 
	A valid argument, therefore, could be that our hybrid model has a more complex structure and more number of parameters than a standard CNN model (the baseline
	in our work), 
	and hence, it could be learning representations
	for images in a better way than a CNN alone.
	But, the results presented in this work should be 
	taken with a pinch of salt and should be understood
	that the GNN in the hybrid model enforces some
	form of structural constraints to be learned along
	with the standard convolutions over the input image
	as is done in CNNs. However, we have made sure that
	percentage difference in the number of parameters between the two models (CNN and the hybrid) is approximately $10\%$.

	
	
	
	
	
	\section{Related Work}
	\label{sec:related_work}
	
	Various graph-based learning approaches for superpixel image classification can be distinguished based on how they perform neighbourhood aggregation. 
	In MoNet~\cite{monti2017geometric}, weighted aggregation is performed by learning a scale factor based on geometric distances. They also test their model on the MNIST Superpixels dataset. 
	In Graph Attention Networks (GATs)~\cite{avelar2020superpixel}, self-attention weights are learned. 
	SplineCNN~\cite{fey2018splinecnn} uses B-spline bases for aggregation, whereas SGCN~\cite{danel2020spatial} is a variant of MoNet and uses a different distance metric. 
	
	There are some preliminary works on enhancing semantic segmentation based systems using superpixel representations. For instance, 
	Mi \textit{et al.}~\cite{mi2020semseg} propose a Superpixel-enhanced Region Module (SRM), which they train jointly with a Deep Neural Forest. The SRM alleviates the noise by leveraging the pixel-superpixel associations. 
	In DrsNet~\cite{yu2021drsnet}, coarse superpixel masking and fine superpixel masking are applied to the CNN features of the input image, particularly for rare classes and background areas.
	Authors of \cite{zhao2017superpixel} have used superpixel-based multiple local regions joint representation CNN model to classify very high resolution (VHR) panchromatic and multispectral (MS) remote-sensing images.
	In \cite{yang2021image}, the authors propose an image recognition method that is based on superpixels and feature fusion, where the features (such as global, texture, appearance features) are computed from the superpixels representation.
	In \cite{liu2019superpixel}, a superpixel-guided layer-wise embedding CNN is devised by the authors, whose main focus is remote sensing image classification. Superpixels guide the network when it comes to unlabelled samples, and superpixels help in handling irregular spatial dependencies.

	\section{Conclusion}
	\label{sec:conclusion}
	Our present work demonstrates that superpixels-based graphs represent domain knowledge for images and that infusing this knowledge in a standard vision-based training process shows significant gains in predictive accuracy. 
	We are able to conclude that a GNN can deal  with relational information conveyed by a superpixel graph and can construct high-level 
	relations that can boost the predictive performance when coupled with a CNN model.
	A straightforward extension is to employ a  pre-trained CNN model for domain-specific tasks and observe how the superpixel information impacts the predictive performance. 
	
	\section*{Data and Code Availability}
	
	Our code repository is available at:  \url{https://github.com/abheesht17/super-pixels}. 
	We provide details on how to obtain the data used in our experiments.
	We also provide
	details on how to use our code-repository for applications in 
	other domains. 
	The authors can be contacted for more details on the code package.
	
	\begin{acks}
		We sincerely thank Jean-Louis Qu\'eguiner, Head of Artificial Intelligence, Data and Quantum Computing at OVHCloud\footnote{\url{https://www.ovhcloud.com/}} for providing us with the necessary computing resources for conducting our experiments.
	\end{acks}
	
	\bibliographystyle{ACM-Reference-Format}
	\bibliography{main}

\end{document}